\documentclass{article}

\usepackage{PRIMEarxiv}

\usepackage[utf8]{inputenc} 
\usepackage[T1]{fontenc}    
\usepackage{hyperref}       
\usepackage{url}            
\usepackage{booktabs}       
\usepackage{amsfonts}       
\usepackage{nicefrac}       
\usepackage{microtype}      
\usepackage{lipsum}
\usepackage{fancyhdr}       
\usepackage{graphicx}       
\graphicspath{{media/}}     

\title{Towards early prediction of neurodevelopmental disorders: Computational model for Face Touch and Self-adaptors in Infants
}

\author{
  Bruno Tafur \\
  University of Cambridge \\
  United Kingdom \\
  \texttt{bt403@cantab.ac.uk} \\
 \And
  Staci Weiss \\
  University of Cambridge \\
  United Kingdom \\
  \texttt{smw95@cam.ac.uk} \\
   \And
  Marwa Mahmoud \\
  University of Glasgow \\
  United Kingdom \\
  \texttt{marwa.mahmoud@glasgow.ac.uk} \\
}

\begin{document}
\maketitle

\begin{abstract}
Infants' neurological development is heavily influenced by their motor skills. Evaluating a baby's movements is key to understanding possible risks of developmental disorders in their growth. Previous research in psychology has shown that measuring specific movements or gestures such as face touches in babies is essential to analyse how babies understand themselves and their context. This research proposes the first automatic approach that detects face touches from video recordings by tracking infants' movements and gestures. The study uses a multimodal feature fusion approach mixing spatial and temporal features and exploits skeleton tracking information to generate more than 170 aggregated features of hand, face and body. This research proposes data-driven machine learning models for the detection and classification of face touch in infants. We used cross dataset testing to evaluate our proposed models. The models achieved 87.0\% accuracy in detecting face touches and 71.4\% macro-average accuracy in detecting specific face touch locations with significant improvements over Zero Rule and uniform random chance baselines. Moreover, we show that when we run our model to extract face touch frequencies of a larger dataset, we can predict the development of fine motor skills during the first 5 months after birth.
\end{abstract}


\section{Introduction}

\begin{figure*}[h]
\centering
\includegraphics[width=0.85\textwidth]{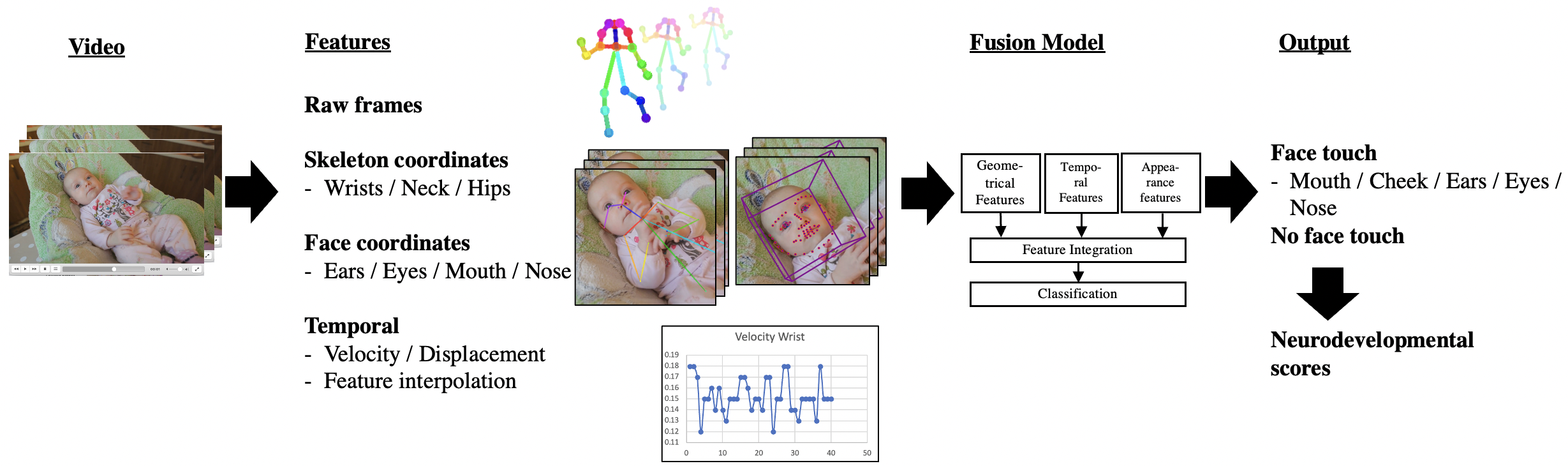}
\caption{An overview of our proposed framework. Spatial, temporal and appearance features are extracted, then they are concatenated with a feature integration layer and a classification approach is used to detect and classify the infant's face touch, which is subsequently used to predict the neurodevelopmental scores.}
\end{figure*}

Analysing body movements in early childhood gives insights into the infant's neurological development, and it can play an essential role in determining if a baby is suffering from injuries in the nervous system or a hereditary disease \cite{akcakaya_correlation_2019}. Also, specific movements such as face touches have been found to be crucial in the development of babies since their fetal age \cite{reissland_goal_2018}. For example, face touches to sensitive areas of the face such as the mouth are frequent in gestational age as babies get prepared to feed. Different cultures have also shown differences in self-touch in babies, and age has also been established as a determinant factor \cite{reissland_goal_2018}.

Various methods are utilised to track and measure movements in babies, including 3D motion capture, sensors and video cameras \cite{marcroft_movement_2015}. 3D motion capture and sensors are mostly used in laboratory settings instead of a more natural setting for the infant as it requires specific equipment and tools \cite{chambers_computer_2020}. There have been few studies that utilised computer vision with video cameras. This method has the advantages of being highly flexible to different environments, giving high contextual information and being easier to interpret \cite{marcroft_movement_2015}. However, it requires more complex computations, depends on the camera quality, angle and movement, and is difficult to generalise to untrained cases.

Various studies have explored the fidgety movements of babies by analysing pixel displacement in video frames \cite{adde_using_2009, adde_identification_2013, adde_early_2010}. Recently, some studies have begun to examine more robust tracking algorithms for body parts based on methods such as OpenPose \cite{chambers_computer_2020,reich_novel_2021}. Despite being an important measure for neurodevelopment, no previous research has looked at specific gesture detection in much detail, such as detecting specific hand movements. Most research is centred on general fidgety movements or general statistics descriptors. Also, these approaches have primarily focused on a general classification for high-risk infants or cerebral palsy \cite{chambers_computer_2020, stoen_computer-based_2017} or classification of movement types \cite{adde_using_2009,reich_novel_2021,das_vision-based_2018}. In the case of face touches, research is limited and has centred on hand-over face gestures in adults \cite{mahmoud_interpreting_2011} or touches from the mother to a child \cite{chen_cnn_2019, chen_touch_2016}.

This research proposes a machine learning model for automatic detection and classification of face touches in newborn infants and their location around key areas of the face using features extracted from raw videos. It proposes using feature selection and fusion models based on temporal and spatial features, using geometric and appearance features of the infant's face and body. The proposed models are validated and evaluated on a couple of datasets and then applied to a large video dataset to extract gesture descriptors of video of one-month-old infants automatically. Using regression, we demonstrate the effectiveness of extracting these gestures in predicting Mullen neurodevelopmental scores \cite{mullen_mullen_1995} for the same infants. To the best of our knowledge, this research represents the first study to analyse specific gestures in infants at this level of granularity and the first to analyse self-touch. The main contributions can be described as follows:

\begin{itemize}
    \item Proposing a data-driven machine learning model for detection and classification of face touch in infants exploiting spatial and temporal features.
    
    \item Evaluating and validating the proposed method in a cross-dataset manner using challenging naturally collected datasets on infants.
    
    \item Presenting preliminary results on using our proposed computational model to detect face touch features in infants on a larger labelled dataset and demonstrating the ability to predict neurodevelopmental scores of the infants using automatic face touch dynamics.
    
    \item Our proposed trained validated model is available on Github as an open source tool for the community. We believe this work will enable future research in infant behaviour modelling and provide a tool for future neurodevelopmental studies.
\end{itemize}

\section{Related Work}

The most relevant studies that use computer vision in the context of neurodevelopment analysis in infants have centered on tracking general movement indicators; such as aggregated data from pose coordinates \cite{chambers_computer_2020, reich_novel_2021} or displacement information from overall images \cite{adde_using_2009, adde_identification_2013, adde_early_2010}. In the analysis of touch, a couple of studies have centred on an analysis of a controlled environment where a mother touches her child \cite{chen_cnn_2019, chen_touch_2016}.

Infants have different proportions in their limbs than adults, making them more complex for general tracking mechanisms to work with the same accuracy. Therefore, the study carried out by Chambers et al. adapted tracking methods based on computer vision on babies \cite{chambers_computer_2020}. Their study focused on developing a tracking method for infants' skeleton coordinates, and they used these statistics to compare the risk of developing neuromotor impairment in healthy and at-risk infants. Their study expanded on OpenPose \cite{cao_openpose_2019} implementation for humans by tuning the model to be used on infant videos.

Regarding more specific movement patterns related to neurological disorders, a study by Das et al. \cite{das_vision-based_2018} focused on analysing the kicking patterns of at-risk infants. Their method tracked their movements using OpenPose and extracted additional KAZE features \cite{alcantarilla_kaze_2012} as image descriptors. In their experiments, the authors used an SVM classifier to differentiate the kicking pattern types as simultaneous movement (SM), non-simultaneous movement (NSM) and no movement (NM).

In the case of touch, Chen et al. performed a couple of studies focused on the detection of interactions between a caregiver and a child in a controlled environment \cite{chen_cnn_2019, chen_touch_2016}. Specifically, their research focused on detecting touch from the caregiver to the child in particular locations: head, arms, legs, hand, torso and feet. Their latest study applied two main methods; firstly, they extracted tracking information by detecting the skeleton locations. Secondly, they extracted the infant's location in the image by applying image segmentation using the GrabCut algorithm.

Therefore, although some studies have attempted to tackle some of these issues in infants, most of them have focused on the analysis of general movements instead of specific gestures.  This study analyses hand to face gestures in infants in larger detail and granularity and proposes novel machine learning models for automatic detection.

The relationship between face and body touch in infants and how they correlate with cognitive development has not been studied quantitatively and systematically before in previous literature. The Mullen Scales of Early Learning is used to measure
the cognitive development of infants in five different categories: gross motor (GM), fine motor (FM), visual reception (VR), receptive language (RL) and expressive language (EL) \cite{mullen_mullen_1995, milosavljevic_adaptation_2019}. They are a key measure of the development of the child during the first years after birth. Previous studies have not tackled the relationship of detected features with MSEL scores. We aim to analyse this relationship based on gesture and movement data extracted from the infant.

\section{Datasets}

For our data-driven models, we used two main datasets: BRIGHT \cite{the_bright_project} and Chambers \cite{chambers_computer_2020}. A subset of the two datasets was labelled and validated by a psychology expert to be later used for our models. Then, the videos from the BRIGHT dataset were used to evaluate the correlations between face touch dynamics and neurodevelopmental scores.
\subsection{BRIGHT dataset} 
This dataset was provided by the 'removed for anonymous submission' and is part of the studies carried out in the Brain Imaging for Global Health (BRIGHT) Project \cite{the_bright_project} in which they study infants from Gambia and UK during their first 24 months of life. The initial sample provided included 29 videos of UK infants. From the 29 videos, 23 videos were selected as some of the babies were occluded during most of the video runtime. Each video shows the behaviour of one infant of fewer than 2 months of age, actively responding to the input given by their mother. The videos were recorded in different rooms, with the infant lying down with a mirror positioned on the wall behind the head of the baby. The camera is static, and the infants generally cover a small portion of the frame but can be located in different parts of the frame. Another complex factor that characterises this dataset includes the mother's presence during the video, sometimes occupying a significant part of the frame with a bigger skeleton and limbs. Also, the fact that the infants are lying down while the camera is facing the front means that the camera generally captures the babies' faces from a side or the bottom, making them difficult to detect for traditional algorithms. The babies are shown rotated in the frames at different angles between 90° and -90°.
    
\subsection{Chambers dataset}
This is an open dataset compiled and generated by Chambers et al \cite{chambers_computer_2020}. 25 videos were selected based on the age of the infants in the video by filtering and selecting only the videos with babies less than 2 months to ensure better consistency with the BRIGHT dataset. The videos show babies lying down on their own and interacting in a natural environment. They could be dancing, playing or rolling over in their crib. The camera is sometimes moving while filming the baby, and the babies generally cover most of the frame. The videos do not feature other people in the frame, but the babies sometimes can move at different angles. Also, the resolutions are very varied between videos, with some of them being more blurry and with smaller frames. The babies are shown rotated in the frames at different angles between 90° and -90°.

\section{Labelling}

The labelling process was carried out using a tagging system developed for this research which allowed efficient tagging of the image frames. Also, the tagging was carried out with the support of a psychology expert, who helped by labelling part of the dataset and providing her judgement about the different labelling categories. 

As this study aims to detect hand over face gestures in infants automatically, the main labelling category to tag needed to differentiate between face touch or no touch in each frame. Therefore, it was defined as follows:
\begin{itemize}
    \item On Head: From a human perspective, it can be seen that the hand could be touching the head area. In this study, the head area considers any of the following locations or any area enclosed by the those locations: eyes, ears, nose, mouth, cheeks, forehead and neck.
    \item Outside Head: From a human perspective, it can be seen that the hand is not on the head area as defined.
\end{itemize}

Additionally, we labelled our dataset with the following non-exclusive categories: eyes, ears, nose, mouth and cheeks, as they are the main differentiable parts of the face. The categories were also discussed and agreed upon with the psychology expert to validate their significance and usefulness from the neurodevelopment perspective. 

The final labelled datasets sizes and distributions can be seen in Table I. The final proportion of ``on head'' versus ``outside head'' was of 29.5\% to 70.5\%, which is expected for this kind of natural dataset.

\begin{table}
\caption{Database sizes and labels}
\begin{center}
\begin{tabular}{|c|c|c|}
\hline
\textbf{Dataset} &\textbf{}  &\textbf{Sizes}  \\
\hline
 & \# Videos & 25 \\
 & Total Frames & 1769 \\
Chambers & Mean Frames per video  & 70.76  \\
&  \% On Head & 29.2\%   \\
&  \% Outside Head & 70.8\%   \\

\hline
 & \# Videos & 23 \\
 & Total Frames & 2039 \\
BRIGHT & Mean Frames per video  & 88.6 \\
&  \% On Head & 29.7\% \\
&  \% Outside Head & 70.3\% \\
\hline
& Total Number of Frames & 3808 \\
\hline
\end{tabular}
\label{results42}
\end{center}
\end{table}

\section{METHOD}

Because of the small size of the labelled dataset, we could not use an end-to-end deep learning model. In this section we present the feature extraction and selection steps and the proposed feature fusion machine learning model.


\subsection{Feature extraction of face and body}
Our proposed models required spatial and temporal features related to the infants' face touch gestures. The features extracted were selected considering the relationship between the hands of the baby and the face. 

\subsubsection{Extraction of face and body landmarks}

We first extracted basic face and body landmarks. 

- Pose coordinates: Positions of the skeleton parts were extracted for every baby and every frame by using the fine-tuned OpenPose \cite{cao_openpose_2019} model trained by Chambers et al. \cite{chambers_computer_2020}. Following the implementation of Chambers et al., the raw pose locations were normalised, smoothed and interpolated per video. 

- Face Region: Based on the extracted pose features and estimated orientation, an accurate estimate of the baby's face location was carried out and the image was cropped in the face region. If no possible face was found in a given frame, the locations of the face of the nearest frames were used as guidance. Where possible, the face region was further aligned based on the locations of the eyes and nose. 

- Face coordinates: OpenPose provides general locations of the eyes, nose and ears, but its purpose is centred on getting the whole skeleton and not on specific facial landmarks. Therefore, information about the location of facial features based on 3D-FAN \cite{bulat2017far} was also used. The faces were extracted from the aligned cropped face regions. 

\begin{figure}[h]
\centering
\includegraphics[width=0.45\textwidth]{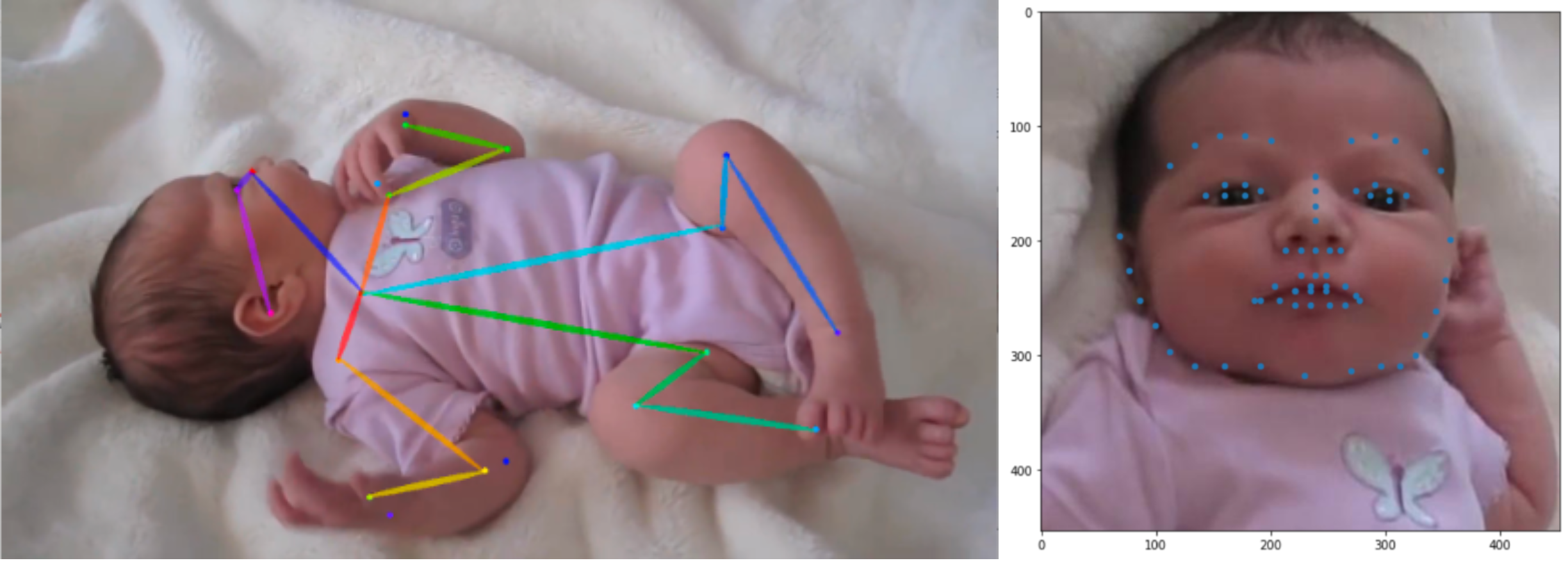}
\caption{Example body skeleton extracted using OpenPose and face keypoints extracted using 3D-FAN}
\end{figure}

\subsubsection{Extraction of geometric, appearance and temporal features}

After basic landmarks features were extracted, we extracted a set of geometric and temporal feature descriptors. 

Based on the initial features, the following features were calculated:

- Face and body geometrical features (Distance and Angular): Based on the coordinates of the skeleton of the baby, the normalised distances between the wrists and the ears, eyes, neck and nose were extracted. For each case, the distances considered included differences in the X direction, differences in the Y direction and euclidean distances.  Additionally, based on the coordinates of the skeleton of the baby, the angles of the elbows and shoulders were extracted. 

- Hands geometrical features (Distance): As the adapted OpenPose model by Chambers et al. \cite{chambers_computer_2020} only generated the skeleton up to the wrists, additional information was obtained by extending the skeleton to the hands. The MediaPipe detection algorithm \cite{zhang_mediapipe_2020} was used in the area surrounding the wrists to obtain the hand coordinates. Based on the coordinates, the normalised distances between the fingers and the eyes and nose were calculated. The distances included differences in the X direction, differences in the Y direction and euclidean distances. Also, confidence scores were considered as additional features based on the confidence of the MediaPipe algorithm detecting each hand. 


- Temporal features: The temporal features were centred on aggregated information over various frames. We calculated features including displacement, speed and acceleration obtained based on the coordinates of the skeleton of the baby for the wrists and elbows.

- Appearance Features: Histogram of Oriented Gradients (HOG) \cite{dalal_histograms_2005} is a method for feature extraction based on the directionality of the gradients in different locations in an image. This method has shown significant success rate in different image detection tasks including detecting faces and expressions \cite{alamsyah_happy_2021, binli_face_2019,mahajan_face_2017} and detecting gestures \cite{prasuhn_hog-based_2014,krisandria_hog-based_2019, feng_static_2013}. These features were extracted only for the main region of interest, which is the face area. Consequently, these features were extracted from the cropped images of the face. Additionally, we wanted to extract more localised spatial information inside the face. Therefore, more granular HOG features were extracted in two specific face areas: one related to the upper region of the face based on the eyes location and another related to the lower region based on the mouth location. Also, confidence scores were considered as additional features based on the average confidence of the landmarks in each region as calculated by 3D-FAN. 

\begin{figure}[h]
  \centering
  \includegraphics[width=0.45\textwidth]{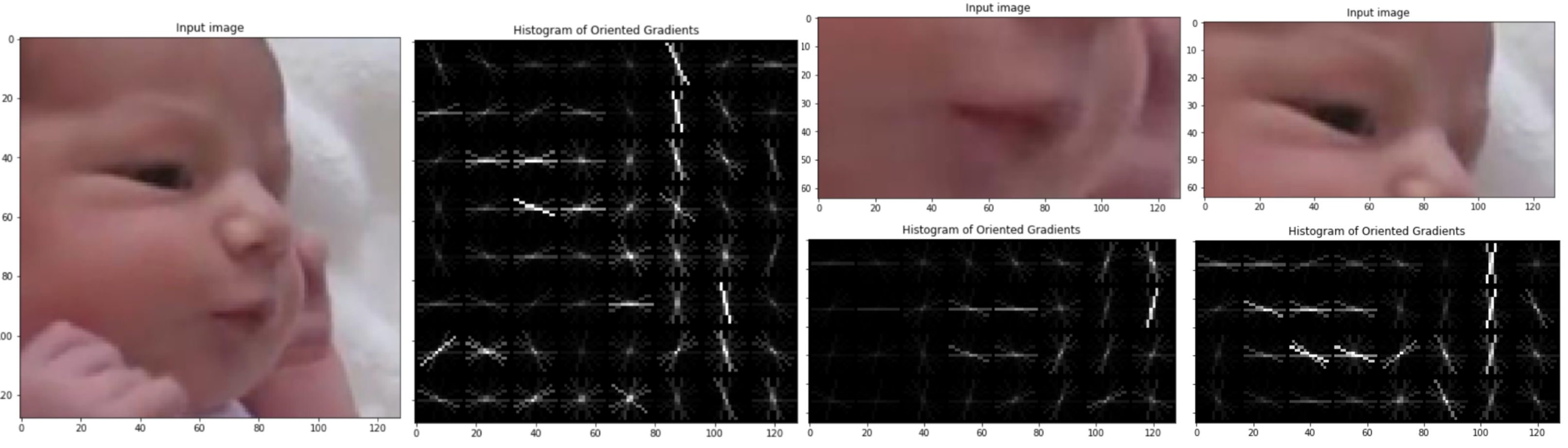}
  \caption{Examples of HOG features obtained for the face region, the upper head and the lower head. Note the challenging nature of the dataset with extreme head poses and viewpoints.}
\end{figure}

\subsubsection{Features smoothing and data augmentation}

As a final step in the feature extraction process, we smoothed and augmented the calculated features to be able to train the classification models on the data. The outliers of the geometrical and temporal features per video were replaced by blank values, and the data was interpolated per video to cover any deleted or missing values. If data was still missing, it was replaced by mean values from the training data during the training stage. 

Finally, to compensate for the small size of the dataset, the training data was augmented by flipping the images horizontally, flipping all the features accordingly and considering the directionality of these features. 

\subsection{Face touch detection and classification}

After feature extraction and smoothing, we handled face touch detection and classification as two different classification problems. The first is to detect when the hand touches the face as a binary classification problem; then, we classify different touch location areas as a multi-label classification problem. The architectures proposed for both problems are very similar. The main difference lies in the method used for the classification component in the final layer. The models can be divided into the following:

\subsubsection{Feature selection and dimensionality reduction}


Our first proposed method used feature selection and dimensionality reduction and a Support Vector Machine (SVM) classifier for solving the face touch detection problem. 

There were four main categories of geometrical and temporal features: body distance features, hand distance features, angular features and temporal features. Many of the features in the same categories correlated with each other as they measured similar characteristics. Therefore, to ensure proper representation of the features, this method proposed reducing these features before training a classifier.

Firstly, the features were filtered based on an automatic feature selection process. Random Forest was used to select the most representative features and prevent skewing the classifier with features that were not that significant. The feature selection was carried out by cross-validating with 5-folds in the training set to ensure independence. Then, Principal Component Analysis (PCA) was used for dimensionality reduction. PCA has shown effective results in detection problems when facing a large number of features \cite{dardas_hand_2011, degen_dimensionality_2015, qi_surface_2020}. PCA was used to filter a percentage of the explained variance. The threshold of this explained variance was established as a hyperparameter that was also learned by cross-validating with 5-folds in the training set. 

After applying PCA, the classification algorithm used SVM using an RBF kernel. The model was cross-validated with 5-folds in the training set to choose the best hyperparameters for SVM. The search for the best hyperparameters for SVM was done in combination with the search for the threshold for PCA, as the hyperparameters were possibly dependent on each other using  a grid search method \cite{pedregosa_scikit-learn_2011}. 

In the case of multi-class classification for the face areas, the Label Powerset model was used with underlying SVMs to predict the multiple overlapping labels. Label Powerset transforms the labels by creating a class for each possible combination of labels and creates a classifier for each combination \cite{read_multi-label_2014}. Consequently, it has the advantage of considering the possible relationships between the labels. This model was configured by tuning the hyperparameters in the same way as in the SVM binary classifier.

\subsubsection{Feature optimisation using deep learned features}


Our second proposed method used autoencoders as a feature optimisation and dimensionality reduction technique. This method has been used in various studies as an effective way of reducing dimensionality while maintaining the representation of the data, and it has been successfully used before with repetitive and correlated features \cite{wang_auto-encoder_2016}. In this case, autoencoders were used to generate a latent representation of the input features. 

Firstly, the dimensions of the features were reduced based on the autoencoder model. The model uses a neural network architecture that learns how to represent the data in lower dimensions and reconstruct it \cite{wang_auto-encoder_2016}. It then minimises the error between the reconstruction and the original input. The aim of the autoencoder is to exploit the correlations in the input features to reduce the final dimensions without losing relevant information. 

This method was used with two alternatives of input features. The first one used only geometrical and temporal features. The second alternative also used the HOG features. The main hyperparameters that were learned for this model included the latent dimensions and the number of epochs. These hyperparameters were selected based on the results of a 5-fold cross-validation in the training set.

After encoding the data, the classification process was done using SVM with an RBF kernel. The input features for the SVM classification process were the output of encoding the features with the trained encoder. The classification layer was also cross-validated with 5-folds in the training set to choose the best hyperparameters for SVM.  Finally, in the case of the multi-label classification problem, the Label Powerset model was used with underlying SVMs to be able to predict the multiple face touch locations.

\section{EVALUATION}

We evaluated the accuracy of the detection of face touches by using a mixture of spatial and temporal features and analysed models based on dimensionality reduction and optimisation techniques. The models were evaluated cross-dataset to validate their effectiveness and generalisation. The approaches were evaluated with three different configurations of the datasets to ensure the consistency of the models. Also, all segmentations of the data were grouped by video to ensure having different videos in each set. The three configurations used were the following:
\begin{itemize}
    \item Train and Cross Validate on BRIGHT dataset - Test on Chambers dataset
    \item Train and Cross Validate on Chambers dataset - Test on BRIGHT dataset
    \item Train and Cross Validate on Chambers and on 50\% of BRIGHT dataset - Test on the other 50\% of BRIGHT dataset 
\end{itemize}

As there was no existing baseline for these models, the models were evaluated against Zero Rule (ZeroR) baseline and random uniform chance. In the case of the ZeroR baseline, it is calculated by assigning the value of the majority class to every data point \cite{de_sa_robust_2020, mahmoud_automatic_2013, bal_performance_2014} while random chance assigns a class based on random uniform probabilities. Statistical McNemar's tests were carried out to ensure the results were significantly different. The McNemar test was used as the compared distributions were binary targets instead of continuous variables. All the best performing models were found significantly different with p $<$ 0.01 in comparison to Random Chance and Zero Rule.

\subsection{Detection of face touches}
\label{predict-face-touch}
The main target was to determine if there was a face touch. This problem was treated as a binary classification task based on the classes: ``on head'' and ``outside head''.

The models that were analysed were the following:
\begin{itemize}
    \item Feature selection and dimensionality reduction based on geometrical and temporal features (RF-PCA-SVM): This model followed the components described in Section 5.2.1. It performed feature selection with Random Forest (RF) and dimensionality reduction with PCA. Finally, it performed the classification of the labels using SVM in the case of this binary problem. It used the geometrical distance and angular features and aggregated temporal features.
    
    \item Feature optimisation using deep learned features based on geometrical and temporal features (AUTO ENC-SVM-I): This model was structured as described in Section 5.2.2. It used an autoencoder neural architecture to reduce the dimensions of the input features. Finally, it performed the classification of the labels using SVM. It used the geometrical features (distance and angular) and the temporal features. 
    
    \item Feature optimisation using deep learned features based on geometrical, temporal and HOG features (AUTOENC-SVM-II): The model was structured as described in Section 5.2.2. Similar to the previous model, it used an autoencoder neural architecture to reduce the dimensions of the input features and SVM for the binary classification problem. It used the geometrical features, temporal features and HOG features.
\end{itemize}

The results for predicting between ``on head'' and ``outside head'' can be seen in Table 2, 3 and 4. All the models had significantly higher accuracy than uniform random chance and ZeroR baselines. The best performing model reached 87\% accuracy when trained in a mixture of both datasets.

Overall the results of the three models were promising with high accuracy in comparison to the baselines. Also, the results were relatively similar between the three models. Some performed better on different datasets, but the performance was very competitive between them. All three models obtained better results than ZeroR or Random Chance in accuracy, precision and recall. Therefore, the results demonstrated that these models can perform well in the detection of face touch. 

Even though the autoencoder models (AUTOENC-SVM) outperformed the random forest and PCA model (RF-PCA-SVM) in two of the three dataset configurations, the difference in accuracy performance was limited. These results demonstrate that the RF-PCA-SVM configuration was also very effective. Possibly in larger datasets, the autoencoder based models could extract more representative features that could better outperform the RF-PCA-SVM model. 

Similarly, the inclusion of the HOG features in the AUTOENC-SVM-II model did not show a noticeable increase in performance. In the case of the BRIGHT dataset, it did show an improvement over the other models and a higher improvement over AUTOENC-SVM-I. However, the improvement could have been greater. This could be caused by the limited amount of data with very varied head poses and rotations. Therefore, the AUTOENC-SVM-II model might perform better if trained in larger datasets where the HOG features can be learned with more generalisable representations. 

Finally, even though there were various challenges in the datasets that could have a negative impact on the models' ability to generalise between datasets, the results demonstrated that the proposed methods had high performance in the detection of face touches.

\begin{table}
\caption{Results -binary classification ``on head'' vs ``outside head''. \\Training and CV dataset: Chambers. Testing dataset: Bright.}
\begin{tabular}{|c|c|c|c|}
\hline
\textbf{Model}   &\textbf{Accuracy} &\textbf{Precision} &\textbf{Recall} \\
\textbf{}   &\textbf{Test} &\textbf{On Head} &\textbf{On Head}\\
\hline
Random Chance & 50\% & 29.7\% & 50\% \\
Zero Rule & 70.3\% & 0\% & 0\% \\
RF-PCA-SVM & 80.3\% & 68.7\% & 62.2\% \\
\textbf{AUTOENC-SVM-I} & \textbf{80.7\%} & \textbf{70.8\%} & \textbf{59.6\%} \\
AUTOENC-SVM-II & 80.6\% & 74.4\% & 53.1\%\\
\hline
\end{tabular}
\label{results12}
\end{table}

\begin{table}
\caption{Results -binary classification ``on head'' vs ``outside head''. \\Training and CV dataset: Bright. Testing dataset: Chambers.}
\begin{tabular}{|c|c|c|c|}
\hline
\textbf{Model}   &\textbf{Accuracy} &\textbf{Precision} &\textbf{Recall}  \\
\textbf{}   &\textbf{Test} &\textbf{On Head} &\textbf{On Head}  \\
\hline
 Random Chance & 50\% & 29.2\% & 50\% \\
 Zero Rule & 70.8\% & 0\% & 0\% \\
 RF-PCA-SVM& 77.8\% & 58.8\% & 80.2\%\\
AUTOENC-SVM-I & 75.2\% & 57.9\% & 54.8\%\\
\textbf{AUTOENC-SVM-II} & \textbf{79.6\%} & \textbf{65.4\%} & \textbf{63.8\%}\\
\hline
\end{tabular}
\label{results13}
\end{table}

\begin{table}
\caption{Results -binary classification ``on head'' vs ``outside head''. \\Training and CV dataset: Chambers + 50\% Bright. Testing dataset: 50\% Bright.}
\begin{tabular}{|c|c|c|c|c|}
\hline
\textbf{Model}   &\textbf{Accuracy} &\textbf{Precision} &\textbf{Recall}  \\
\textbf{}   &\textbf{Test} &\textbf{On Head} &\textbf{On Head} \\
\hline
Random Chance & 50\% & 28.3\% & 50\% \\
Zero Rule & 71.7\% & 0\% & 0\% \\
\textbf{RF-PCA-SVM} & \textbf{87.0\%} & \textbf{77.2\%} & \textbf{76.8\%} \\
 AUTOENC-SVM-I & 86.9\% & 76.9\% & 77.1\% \\
AUTOENC-SVM-II & 85.7\% & 71.6\% & 82.2\% \\
\hline
\end{tabular}
\label{results15}
\end{table}

\subsection{Classification of face touch descriptors}

These experiments evaluate the face touch on specific locations of the face. These locations were evaluated based on the universe of images where there is a face touch. The key locations to predict included the following: ears, nose, cheeks, mouth, and eyes. 
The problem mas evaluated as a multi-label problem because the different classes could overlap and the infant could touch more than one location at the same time.

The proposed models for this problem are the same as the ones described in Section \ref{predict-face-touch}, so we will use the same naming abbreviations. The main difference was the change in the classification method from SVM to Label Powerset with SVM \cite{read_multi-label_2014} to tackle the problem as a multi-label classification problem. Therefore, the models that were analysed were the following:
\begin{itemize}
    \item Feature selection and dimensionality reduction based on geometrical and temporal features (RF-PCA-SVM)
    \item Feature optimisation using deep learned features based on geometrical and temporal features (AUTOENC-SVM-I)
    \item Feature optimisation using deep learned features based on geometrical, temporal and HOG features (AUTOENC-SVM-II)
\end{itemize}

The experiments were carried out only on the portion of images labelled as ``on head'' so that it could be sufficiently balanced; therefore the dataset was even more limited in size than the original.

The obtained results can be seen in Table 5, 6 and 7. The results show the macro-average accuracy, precision and recall of the multiple key locations per model. The highest performing model reached 71.4\% average accuracy when testing on the Chamber's dataset. 

\begin{table}
\caption{Results of predicting key areas\\Training and CV dataset: Chambers. Testing dataset: Bright.}
\begin{tabular}{|c|c|c|c|}
\hline
\textbf{Model}   &\textbf{Accuracy} &\textbf{Precision} &\textbf{Recall} \\
\textbf{}   &\textbf{Test} &\textbf{Key Area} &\textbf{Key Area}  \\
\hline
 Random Chance & 50.0\% & 31.1\% & 50.0\%\\
Zero Rule & 24.5\% & 13\% & 0\%\\
\textbf{RF-PCA-SVM} &  \textbf{66.6\%} &  \textbf{33.5\%} &  \textbf{43.8\%} \\
AUTOENC-SVM-I & 63.2\% & 49.6\% & 16.7\%\\
AUTOENC-SVM-II & 63.0\% & 60.9\% & 13.2\% \\
\hline
\end{tabular}
\label{results21}
\end{table}

\begin{table}
\caption{Results of predicting key areas\\Training and CV dataset: Bright. Testing dataset: Chambers.}
\begin{tabular}{|c|c|c|c|}
\hline
\textbf{Model}   &\textbf{Accuracy} &\textbf{Precision} &\textbf{Recall} \\
\textbf{}   &\textbf{Test} &\textbf{Key Area} &\textbf{Key Area} \\
\hline
Random Chance & 50.0\% & 20.8\% & 50\% \\
Zero Rule & 37.8\% & 15\% & 0\% \\
 RF-PCA-SVM & 62.5\% & 36.3\% & 18.8\% \\
AUTOENC-SVM-I & 63.8\% & 29.7\% & 34.3\% \\
\textbf{AUTOENC-SVM-II} & \textbf{71.4\%} & \textbf{35.7\%} & \textbf{24.9\%}\\
\hline
\end{tabular}
\label{results22}
\end{table}

\begin{table}
\caption{Results of predicting key areas\\Training and cross-validation dataset: Chambers + 50\% Bright. Testing dataset: 50\% Bright.}
\begin{tabular}{|c|c|c|c|}
\hline
\textbf{Model}   &\textbf{Accuracy} &\textbf{Precision} &\textbf{Recall} \\
\textbf{}   &\textbf{Test} &\textbf{Key Area} &\textbf{Key Area} \\
\hline
Random Chance & 50.0\% & 29.1\% & 50.0\%\\
Zero Rule & 21.8\% & 17.0\% & 0\%\\
RF-PCA-SVM & 60.3\% & 56.0\% & 34.3\%\\
AUTOENC-SVM-I & 58.1\% & 48.2\% & 19.2\%\\
\textbf{AUTOENC-SVM-II} &  \textbf{60.7\%} & \textbf{44.3\%} & \textbf{16.7\%}\\
\hline
\end{tabular}
\label{results23}
\end{table}

The main metric established to select the best models during cross-validation was the average macro-accuracy of the locations; so this was used as the main indicator of the models performance. The results demonstrated that the models performed effectively better than the baselines. 

As expected, the accuracies were lower than for the face touch problem as this was a complex multi-label problem where multiple locations can overlap on the face, and it can be difficult even for a human to determine the exact location.

Likewise to the previous task, the results were similar between models, but these results show some indication that HOG features might be useful in some instances. The AUTOENCODER-SVM-II model outperformed the accuracies of the other models in two cases and demonstrated a considerable difference in accuracy when it was trained in the BRIGHT dataset. Possibly training with HOG features in more extensive and more varied datasets could make their representations more stable and significant in the end results. 

\subsection{Predicting neurodevelopment scores}
The next step was to evaluate our proposed model results - detected face touch dynamics of infants less than 2 months old - on predicting their neurodevelopmental rates collected at ages 3 and 5 months. We chose the best-performing model for the binary classification task (RF-PCA-SVM) and ran it on a larger dataset. Since we had the Mullen scores only for the BRIGHT dataset, we ran our model on an average of 490 frames per video (19 videos), a total of 9298 frames. We then extracted the face touch frequency for each infant and evaluated it versus the Mullen Scales of Early Learning (MSEL) related to gross motor (GM) skills and fine motor (FM) skills. In this case, the data was limited because the provided metrics were evaluated per infant, and only 19 infants of the BRIGHT dataset had their information available. 

In the case of the MSEL metrics, the data consisted of raw scores per visit of the infant related to the different MSEL categories. A rate of development was calculated per infant per category based on the rate of increase during their first five months. The data used for this case were the gross motor (GM) skills and the fine motor (FM) skills, as they are related to the infant's motor development and could be related to face touch behaviour. After calculating the rate of development of the GM and FM skills, a correlation was calculated between the ratio of face touches per frame and these rates of development of each child.

The results showed a low to moderate positive correlation between the ratio of face touches and the rate of development during the first months. The correlation coefficient obtained for FM was 0.599 with a significant p-value of 0.0067. The correlation coefficient for GM was 0.186, but the p-value was not found to be significant. It is possible that face touches are more related to fine motor skills as they are more specific and localised movements.

The results indicate that measuring infants' face touch frequencies and dynamics in their first month or two can be a predictive measure of their neurodevelopmental scores. It also demonstrates the effectiveness of our proposed computational model as a tool for the early prediction of neurodevelopmental factors. We make the trained model available to the research community at 'removed for anonymous submission' as a baseline for infant's face touch detection and to facilitate future research in this area on more extensive datasets.

A dataset of 19 infants is limited, so it was not possible to test more complex prediction algorithms. However, these results show that, in more extensive datasets, the face touch frequency could be used as one independent variable to help predict infant neurodevelopment scores such as MSEL. Also, the models proposed during this research could support the automation of the extraction of these face touches.

\section{Conclusion}

Our research proposed a machine learning model for automatic detection of face touches in infants using features extracted from videos. This is the first study to provide a computational model for detection and classification of these types of gestures in infants. Our proposed models using a mix of spatial and temporal features with deep learning features demonstrated significantly high accuracies in predicting face touch and their locations around keypoints in the face establishing a promising step for future research in this area. We also showed the effectiveness of the proposed model in predicting MSEL scores related to fine motor (FM) skills , demonstrating that our proposed model can be used as en early prediction tool for neurodevelopmental disorders in infants and it is considered a baseline for future work in this domain. We believe this research will open the door for future research in this area both on the technical as well as neurodevelopmetanl psychology fronts.
 
Despite the promising results, there are several limitations to our model. The datasets used were recorded in almost uncontrolled environments, with varied camera angles and the mum's presence in most videos. These characteristics made the labelling as well as the classification tasks very challenging. We are also aware of the small size of the datasets used in this work. Obtaining datasets, especially for infants, is a challenging task due to the privacy and ethical factors that need to be considered. However, we believe this research serves as a baseline for infant face touch detection and classification and will open the door for further research on more extensive datasets in this area. 

\bibliographystyle{unsrt}  
\bibliography{references}

\end{document}